# Studying Satellite Image Quality Based on the Fusion Techniques

Firouz Abdullah Al-Wassai[*]
Research Student,
Computer Science Dept.
(SRTMU), Nanded, India
fairozwaseai@yahoo.com

N.V. Kalyankar
Principal,
Yeshwant Mahavidyala College
Nanded, India
drkalyankarnv@yahoo.com

Ali A. Al-Zaky
Assistant Professor,
Dept. of Physics, College of Science,
Mustansiriyah Un., Baghdad – Iraq
dr.alialzuky@yahoo.com

*Abstract*: Various and different methods can be used to produce high-resolution multispectral images from high-resolution panchromatic image (PAN) and low-resolution multispectral images (MS), mostly on the pixel level. However, the jury is still out on the benefits of a fused image compared to its original images. There is also a lack of measures for assessing the objective quality of the spatial resolution for the fusion methods. Therefore, an objective quality of the spatial resolution assessment for fusion images is required. So, this study attempts to develop a new qualitative assessment to evaluate the spatial quality of the pan sharpened images by many spatial quality metrics. Also, this paper deals with a comparison of various image fusion techniques based on pixel and feature fusion techniques.

## I. INTRODUCTION:

Image fusion is a process, which creates a new image representing combined information composed from two or more source images. Generally, one aims to preserve as much source information as possible in the fused image with the expectation that performance with the fused image will be better than, or at least as good as, performance with the source images [1]. Image fusion is only an introductory stage to another task, e.g. human monitoring and classification. Therefore, the performance of the fusion algorithm must be measured in terms of improvement or image quality. Several authors describe different spatial and spectral quality analysis techniques of the fused images. Some of them enable subjective, the others objective, numerical definition of spatial or spectral quality of the fused data [2-5]. The evaluation of the spatial quality of the pan-sharpened images is equally important since the goal is to retain the high spatial resolution of the PAN image. A survey of the pan sharpening literature revealed there were very few papers that evaluated the spatial quality of the pan-sharpened imagery [6]. Consequently, there are very few spatial quality metrics found in the literatures. However, the jury is still out on the benefits of a fused image compared to its original images. There is also a lack of measures for assessing the objective quality of the spatial resolution of the fusion methods. Therefore, an objective quality of the spatial resolution assessment for fusion images is required.

Therefore, this study presented a new approach to assess the spatial quality of a fused image based on High pass Division Index (HPDI). In addition, many spectral quality metrics, to compare the properties of fused images and their ability to preserve the similarity with respect to the original MS image while incorporating the spatial resolution of the PAN image, should increase the spectral fidelity while retaining the spatial resolution of the PAN). They take into account local measurements to estimate how well the important information in the source images is represented by the fused image. In addition, this study focuses on cambering that the best methods based on pixel fusion techniques (see section 2) are those with the fallowing feature fusion techniques: Segment Fusion (SF), Principal Component Analysis based Feature Fusion (PCA) and Edge Fusion (EF) in [7].

The paper organized as follows .Section II gives the image fusion techniques; Section III includes the quality of evaluation of the fused images; Section IV covers the experimental results and analysis then subsequently followed by the conclusion.

## II. IMAGE FUSION TECHNIQUES

Image fusion techniques can be divided into three levels, namely: pixel level, feature level and decision level of representation [8-10]. The image fusion techniques based on pixel can be grouped into several techniques depending on the tools or the processing methods for image fusion procedure. In this work proposed categorization scheme of image fusion techniques Pixel based image fusion methods summarized as the fallowing:

a. Arithmetic Combination techniques: such as Bovey Transform (BT) [11-13]; Color Normalized Transformation (CN) [14, 15]; Multiplicative Method (MLT) [17, 18].
b. Component Substitution fusion techniques: such as HIS, HIS, HSV, HLS and YIQ in [19].
c. Frequency Filtering Methods :such as in [20] High-Pass Filter Additive Method (HPFA) , High – Frequency- Addition Method (HFA) , High Frequency Modulation Method (HFM) and The Wavelet transform-based fusion method (WT).
d. Statistical Methods: such as in [21] Local Mean Matching (LMM), Local Mean and Variance

Matching (LMVM), Regression variable substitution (RVS), and Local Correlation Modeling (LCM).

All the above techniques employed in our previous studies [19-21]. Therefore, the best method for each group selected in this study as the fallowing:

a. Arithmetic and Frequency Filtering techniques are High –Frequency- Addition Method (HFA) and High Frequency Modulation Method (HFM) [20].
b. The Statistical Methods it was with Regression variable substitution (RVS) [21].
c. In the Component Substitution fusion techniques the IHS method by [22] it was much better than the others methods [19].

To explain the algorithms through this study, Pixels should have the same spatial resolution from two different sources that are manipulated to obtain the resultant image. Here, The PAN images have a different spatial resolution from that of the original multispectral MS images. Therefore, resampling of MS images to the spatial resolution of PAN is an essential step in some fusion methods to bring the MS images to the same size of PAN, thus the resampled MS images will be noted by $M_k$ that represents the set of DN of band $k$ in the resampled MS image.

## III. QUALITY EVALUATION OF THE FUSED IMAGES

This section describes the various spatial and spectral quality metrics used to evaluate them. The spectral fidelity of the fused images with respect to the original multispectral images is described. When analyzing the spectral quality of the fused images we compare spectral characteristics of images obtained from the different methods, with the spectral characteristics of resampled original multispectral images. Since the goal is to preserve the radiometry of the original MS images, any metric used must measure the amount of change in DN values in the pan-sharpened image $F_k$ compared to the original image $M_k$. Also, In order to evaluate the spatial properties of the fused images, a panchromatic image and intensity image of the fused image have to be compared since the goal is to retain the high spatial resolution of the PAN image. In the following $F_k$, $M_k$ are the measurements of each the brightness values pixels of the result image and the original MS image of band $k$, $\overline{M}_k$ and $\overline{F}_k$ are the mean brightness values of both images and are of size $n*m$. BV is the brightness value of image data $\overline{M}_k$ and $\overline{F}_k$.

### A. Spectral Quality Metrics:

a. **Standard Deviation (SD):** The standard deviation (SD), which is the square root of variance, reflects the spread in the data. Thus, a high contrast image will have a larger variance, and a low contrast image will have a low variance. It indicates the closeness of the fused image to the original MS image at a pixel level. The ideal value is zero.

$$\sigma = \sqrt{\frac{\sum_{i=1}^{m}\sum_{j=1}^{n}(BV(n,m)-\mu)^2}{m \times n}} \quad (1)$$

b. **Entropy ($En$):** The entropy of an image is a measure of information content but has not been used to assess the effects of information change in fused images. En reflects the capacity of the information carried by images. The larger En mean high information in the image [6]. By applying Shannon's entropy in evaluation the information content of an image, the formula is modified as [23]:

$$En = -\sum_{i=0}^{255} P(i)\log_2 P(i) \quad (2)$$

Where P(i) is the ratio of the number of the pixels with gray value equal to $i$ over the total number of the pixels.

c. **Signal-to Noise Ratio (SNR):** The signal is the information content of the data of original MS image $M_k$, while the merging $F_k$ can cause the noise, as error that is added to the signal. The $RMS_k$ of the signal-to-noise ratio can be used to calculate the signal-to-noise ratio $SNR_k$, given by [24]:

$$SNR_k = \sqrt{\frac{\sum_i^n \sum_j^m (F_k(i,j))^2}{\sum_i^n \sum_j^m (F_k(i,j)-M_k(i,j))^2}} \quad (3)$$

d. **Deviation Index (DI):** In order to assess the quality of the merged product in regard of spectral information content. The deviation index is useful parameter as defined by [25,26], measuring the normalized global absolute difference of the fused image $F_k$ with the original MS image $M_k$ as follows:

$$DI_k = \frac{1}{nm}\sum_i^n \sum_j^m \frac{|F_k(i,j)-M_k(i,j)|}{M_k(i,j)} \quad (4)$$

e. **Correlation Coefficient (CC):** The correlation coefficient measures the closeness or similarity between two images. It can vary between –1 to +1. A value close to +1 indicates that the two images are very similar, while a value close to –1 indicates that they are highly dissimilar. The formula to compute the correlation between $F_k$, $M_k$:

$$CC = \frac{\sum_i^n \sum_j^m (F_k(i,j)-\overline{F}_k)(M_k(i,j)-\overline{M}_k)}{\sqrt{\sum_i^n \sum_j^m (F_k(i,j)-\overline{F}_k)^2}\sqrt{\sum_i^n \sum_j^m (M_k(i,j)-\overline{M}_k)^2}} \quad (5)$$

Since the pan-sharpened image larger (more pixels) than the original MS image it is not possible to compute the correlation or apply any other mathematical operation between them. Thus, the upsampled MS image $M_k$ is used for this comparison.

f. **Normalization Root Mean Square Error (NRMSE):** the NRMSE used in order to assess the effects of information changing for the fused image. When level of information loss can be expressed as a function of the original MS pixel $M_k$ and the fused pixel $F_k$, by using the NRMSE between $M_k$ and $F_k$ images in band k. The Normalized Root- Mean-Square Error $NRMSE_k$ between $F_k$ and $M_k$ is a point analysis in multispectral space representing the amount of change the original MS pixel and the corresponding output pixels using the following equation [27]:

$$NRMSE_k = \sqrt{\frac{1}{nm*255^2}\sum_i^n \sum_j^m (F_k(i,j) - M_k(i,j))^2} \quad (6)$$

### B. Spatial Quality Metrics:

a. **Mean Grades (MG):** MG has been used as a measure of image sharpness by [27, 28]. The gradient at any

pixel is the derivative of the DN values of neighboring pixels. Generally, sharper images have higher gradient values. Thus, any image fusion method should result in increased gradient values because this process makes the images sharper compared to the low-resolution image. The gradient defines the contrast between the details variation of pattern on the image and the clarity of the image [5]. MG is the index to reflect the expression ability of the little detail contrast and texture variation, and the definition of the image. The calculation formula is [6]:

$$\bar{G} = \frac{1}{(m-1)(n-1)} \sum_{i=1}^{m-1} \sum_{j=1}^{n-1} \sqrt{\frac{\Delta I_x^2 + \Delta I_y^2}{2}} \quad (7)$$

Where

$$\Delta I_x = f(i+1,j) - f(i,j)$$
$$\Delta I_y = f(i,j+1) - f(i,j) \quad (8)$$

Where $\Delta I_x$ and $\Delta I_y$ are the horizontal and vertical gradients per pixel of the image fused $f(i,j)$. generally, the larger $\bar{G}$, the more the hierarchy, and the more definite the fused image.

**b. Soble Grades (SG):** this approach developed in this study by used the Soble operator is A better edge estimator than the mean gradient. That by computes discrete gradient in the horizontal and vertical directions *at* the pixel location $i,j$ of an image $f(i,j)$. The Soble operator was the most popular edge detection operator until the development of edge detection techniques with a theoretical basis. It proved popular because it gave a better performance contemporaneous edge detection operator than other such as the Prewitt operator [30]. For this, which is clearly more costly to evaluate, the orthogonal components of gradient as the following [31]:

$$Gx = \{f(i-1,j+1) + 2f(i-1,j) + f(i-1,j-1)\} - \{f(i+1,j+1) + 2f(i+1,j) + f(i+1,j-1)\}$$
And
$$Gy = \{f(i-1,j+1) + 2f(i,j+1) + f(i+1,j+1)\} - \{f(i-1,j-1) + 2f(i,j-1) + f(i+1,j-1)\}$$
(9)

It can be seen that the Soble operator is equivalent to simultaneous application of the templates as the following [32]:

$$G_x = \begin{bmatrix} 1 & 2 & 1 \\ 0 & 0 & 0 \\ -1 & -2 & -1 \end{bmatrix} \quad G_y = \begin{bmatrix} -1 & 0 & 1 \\ -2 & 0 & 2 \\ -1 & 0 & 1 \end{bmatrix} \quad (10)$$

Then the discrete gradient $G$ of an image $f(i,j)$ is given by

$$\bar{G} = \frac{1}{(m-1)(n-1)} \sum_{i=1}^{(m-1)} \sum_{j=1}^{(n-1)} \sqrt{\frac{G_x^2 + G_y^2}{2}} \quad (11)$$

Where $G_x$ and $G_y$ are the horizontal and vertical gradients per pixel. Generally, the larger values for $\bar{G}$, the more the hierarchy and the more definite the fused image.

*C. Filtered Correlation Coefficients (FCC):*

This approach was introduced [33]. In the Zhou's approach, the correlation coefficients between the high-pass filtered fused PAN and TM images and the high-pass filtered PAN image are taken as an index of the spatial quality. The high-pass filter is known as a Laplacian filter as illustrated in eq. (12):

$$\text{mask} = \begin{bmatrix} -1 & -1 & -1 \\ -1 & 8 & -1 \\ -1 & -1 & -1 \end{bmatrix} \quad (12)$$

However, the magnitude of the edges does not necessarily have to coincide, which is the reason why Zhou et al proposed to look at their correlation coefficients [33]. So, in this method the average correlation coefficient of the faltered PAN image and all faltered bands is calculated to obtain FCC. An FCC value close to one indicates high spatial quality.

*D. High Pass Deviation Index (HPDI)*

This approach proposed by [25, 26] as the measuring of the normalized global absolute difference for spectral quantity for the fused image $F_k$ with the original MS image $M_k$. This study developed that is quality metric to measure the amount of edge information from the PAN image is transferred into the fused images by used the high-pass filter (eq. 12). which that the high-pass filtered PAN image are taken as an index of the spatial quality. The HPDI wants to extract the high frequency components of the PAN image and each $F_k$ band. The deviation index between the high pass filtered $P$ and the fused $F_k$ images would indicate how much spatial information from the PAN image has been incorporated into the $MS$ image to obtain HPDI as follows:

$$HPDI_k = \frac{1}{nm} \sum_i^n \sum_j^m \frac{|F_k(i,j) - P(i,j)|}{P(i,j)} \quad (13)$$

The smaller value HPDI the better image quality. Indicates that the fusion result it has a high spatial resolution quality of the image.

## IV. EXPERIMENTAL RESULTS

The above assessment techniques are tested on fusion of Indian IRS-1C PAN of the 5.8- m resolution panchromatic band and the Landsat TM the red (0.63 - 0.69 µm), green (0.52 - 0.60 µm) and blue (0.45 - 0.52 µm) bands of the 30 m resolution multispectral image were used in this work. Fig.1 shows the IRS-1C PAN and multispectral TM images. Hence, this work is an attempt to study the quality of the images fused from different sensors with various characteristics. The size of the PAN is 600 * 525 pixels at 6 bits per pixel and the size of the original multispectral is 120 * 105 pixels at 8 bits per pixel, but this is upsampled by nearest neighbor to same size the PAN image. The pairs of images were geometrically registered to each other. The HFA, HFM, HIS, RVS, PCA, EF, and SF methods are employed to fuse IRS-C PAN and TM multi-spectral images. The original MS and PAN are shown in (Fig. 1).

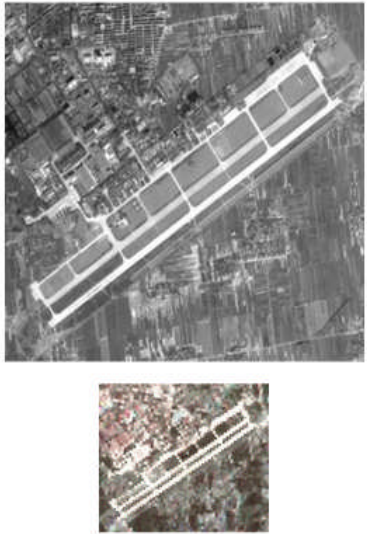

Fig.1: The Representation of Original Panchromatic and Multispectral Images

## V. ANALYSISES RESULTS

### A. *Spectral Quality Metrics Results:*

From table1 and Fig. 2 shows those parameters for the fused images using various methods. It can be seen that from Fig. 2a and table1 the SD results of the fused images remains constant for all methods except the IHS. According to the computation results En in table1, the increased En indicates the change in quantity of information content for spectral resolution through the merging. From table1 and Fig.2b, it is obvious that En of the fused images have been changed when compared to the original MS except the PCA. In Fig.2c and table1 the maximum correlation values was for PCA. In Fig.2d and table1 the maximum results of SNR were with the SF, and HFA.

Results of the SNR, NRMSE and DI appear changing significantly. It can be observed from table1 with the diagram Fig. 2d & Fig. 5e for results SNR, NRMSE & DI of the fused image, the SF and HFA methods gives the best results with respect to the other methods. Means that this method maintains most of information spectral content of the original MS data set which gets the same values presented the lowest value of the NRMSE and DI as well as the high of the CC and SNR. Hence, the SF and HFA fused images for preservation of the spectral resolution original MS image much better techniques than the other methods.

Table 1: The Spectral Quality Metrics Results for the Original MS and Fused Image Methods

| Method | Band | SD | En | SNR | NRMSE | DI | CC |
|---|---|---|---|---|---|---|---|
| ORG | R | 51.018 | 5.2093 | | | | |
| | G | 51.477 | 5.2263 | | | | |
| | B | 51.983 | 5.2326 | | | | |
| EF | R | 55.184 | 6.0196 | 6.531 | 0.095 | 0.138 | 0.896 |
| | G | 55.792 | 6.0415 | 6.139 | 0.096 | 0.151 | 0.896 |
| | B | 56.308 | 6.0423 | 5.81 | 0.097 | 0.165 | 0.898 |
| HFA | R | 52.793 | 5.7651 | 9.05 | 0.068 | 0.08 | 0.943 |
| | G | 53.57 | 5.7833 | 8.466 | 0.07 | 0.087 | 0.943 |
| | B | 54.498 | 5.7915 | 7.9 | 0.071 | 0.095 | 0.943 |
| HFM | R | 52.76 | 5.9259 | 8.399 | 0.073 | 0.082 | 0.934 |
| | G | 53.343 | 5.8979 | 8.286 | 0.071 | 0.084 | 0.94 |
| | B | 54.136 | 5.8721 | 8.073 | 0.069 | 0.086 | 0.945 |
| HIS | R | 41.164 | 7.264 | 6.583 | 0.088 | 0.104 | 0.915 |
| | G | 41.986 | 7.293 | 6.4 | 0.086 | 0.114 | 0.917 |
| | B | 42.709 | 7.264 | 5.811 | 0.088 | 0.122 | 0.917 |
| PCA | R | 47.875 | 5.1968 | 6.735 | 0.105 | 0.199 | 0.984 |
| | G | 49.313 | 5.2485 | 6.277 | 0.108 | 0.222 | 0.985 |
| | B | 51.092 | 5.2941 | 5.953 | 0.109 | 0.245 | 0.986 |
| RVS | R | 51.323 | 5.8841 | 7.855 | 0.078 | 0.085 | 0.924 |
| | G | 51.769 | 5.8475 | 7.813 | 0.074 | 0.086 | 0.932 |
| | B | 52.374 | 5.8166 | 7.669 | 0.071 | 0.088 | 0.938 |
| SF | R | 51.603 | 5.687 | 9.221 | 0.067 | 0.09 | 0.944 |
| | G | 52.207 | 5.7047 | 8.677 | 0.067 | 0.098 | 0.944 |
| | B | 53.028 | 5.7123 | 8.144 | 0.068 | 0.108 | 0.945 |

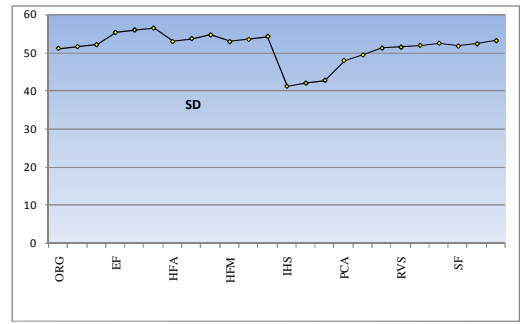

Fig. 2a: Chart Representation of SD

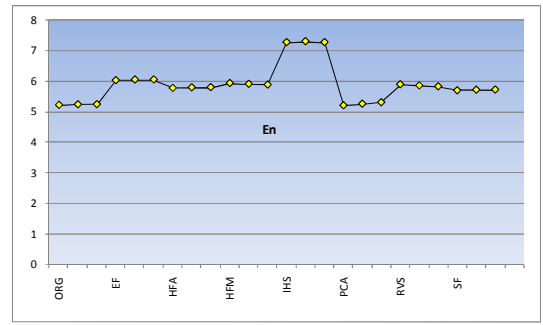

Fig. 2b: Chart Representation of En

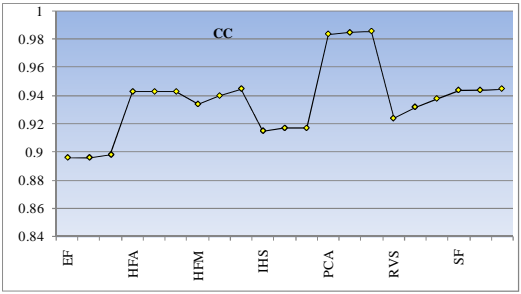

Fig.2c: Chart Representation of CC

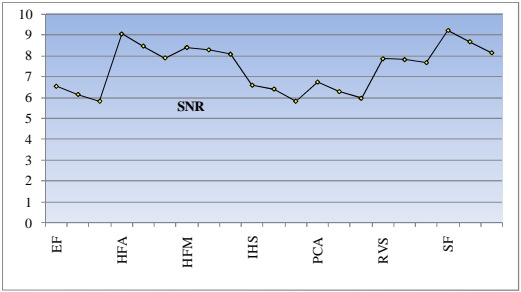

Fig. 2d: Chart Representation of SNR

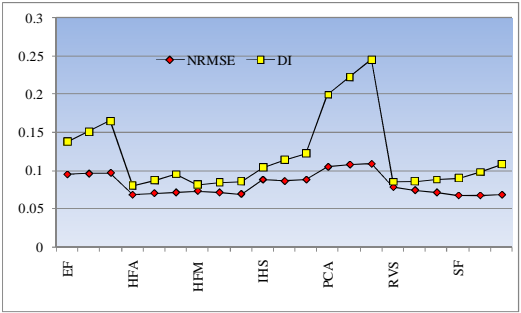

Fig. 2e: Chart Representation of NRMSE&DI

Fig. 2: Chart Representation of SD, En, CC, SNR, NRMSE & DI of Fused Images

### B. Spatial Quality Metrics Results:

Table 2 and Fig. 4 show the result of the fused images using various methods. It is clearly that the seven fusion methods are capable of improving the spatial resolution with respect to the original MS image. From table2 and Fig. 3 shows those parameters for the fused images using various methods. It can be seen that from Fig. 3a and table2 the MG results of the fused images increase the spatial resolution for all methods except the PCA. from the table2 and Fig.3a the maximum gradient for MG was 25 edge but for SG in table2 and Fig.3b the maximum gradient was 64 edge means that the SG it gave, overall, a better performance than MG to edge detection. In addition, the SG results of the fused images increase the gradient for all methods except the PCA means that the decreasing in gradient that it dose not enhance the spatial quality. The maximum results of MG and SG for sharpen image methods was for the EF as well as the results of the MG and the SG for the HFA and SF methods have the same results approximately. However, the comparing them to the PAN it can be seen that the SF close to the result of the PAN. Other means the SF added the details of the PAN image to the MS image as well as the maximum preservation of the spatial resolution of the PAN.

Table 2: The Spatial Quality Metrics Results for the Original MS and Fused Image Methods

| Method | Band | MG | SG | HPDI | FCC |
|---|---|---|---|---|---|
| **EF** | R | 25 | 64 | 0 | -0.038 |
| | G | 25 | 65 | 0.014 | -0.036 |
| | B | 25 | 65 | 0.013 | -0.035 |
| **HFA** | R | 11 | 51 | -0.032 | 0.209 |
| | G | 12 | 52 | -0.026 | 0.21 |
| | B | 12 | 52 | -0.028 | 0.211 |
| **HFM** | R | 12 | 54 | 0.001 | 0.205 |
| | G | 12 | 54 | 0.013 | 0.204 |
| | B | 12 | 53 | 0.02 | 0.201 |
| **IHS** | R | 9 | 36 | 0.004 | 0.214 |
| | G | 9 | 36 | 0.009 | 0.216 |
| | B | 9 | 36 | 0.005 | 0.217 |
| **PCA** | R | 6 | 33 | -0.027 | 0.07 |
| | G | 6 | 34 | -0.022 | 0.08 |
| | B | 6 | 35 | -0.021 | 0.092 |
| **RVS** | R | 13 | 54 | -0.005 | -0.058 |
| | G | 12 | 53 | 0.001 | -0.054 |
| | B | 12 | 52 | 0.006 | -0.05 |
| **SF** | R | 11 | 48 | -0.035 | 0.202 |
| | G | 11 | 49 | -0.026 | 0.204 |
| | B | 11 | 49 | -0.024 | 0.206 |
| **MS** | R | 6 | 32 | -0.005 | 0.681 |
| | G | 6 | 32 | -0.004 | 0.669 |
| | B | 6 | 33 | -0.004 | 0.657 |
| **PAN** | 1 | 10 | 42 | — | — |

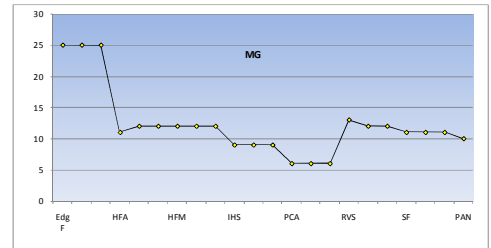

Fig. 3a: Chart Representation of MG

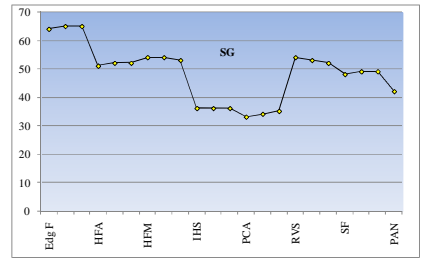

Fig. 3b: Chart Representation of SG

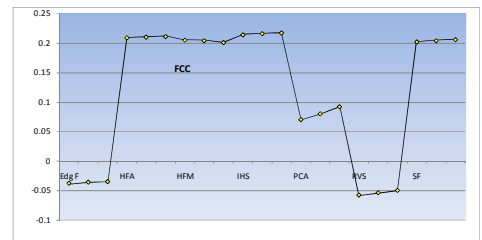

Fig. 3c: Chart Representation of FCC
Continue ⟶

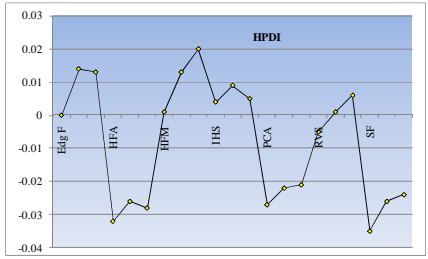

Fig. 3d: Chart Representation of HPDI

Fig. 3: Chart Representation of MG, SG, FCC & HPDI of Fused Images

According to the computation results, FCC in table2 and Fig.2c the increase FCC indicates the amount of edge information from the PAN image transferred into the fused images in quantity of spatial resolution through the merging. The maximum results of FCC From table2 and Fig.2c were for the SF, HFA and HFM. The results of HPDI better than FCC it is appear changing significantly. It can be observed that from Fig.2d and table2 the maximum results of the purpose approach HPDI it was with the SF and HFA methods. The purposed approach of HPDI as the spatial quality metric is more important than the other spatial quality matrices to distinguish the best spatial enhancement through the merging.

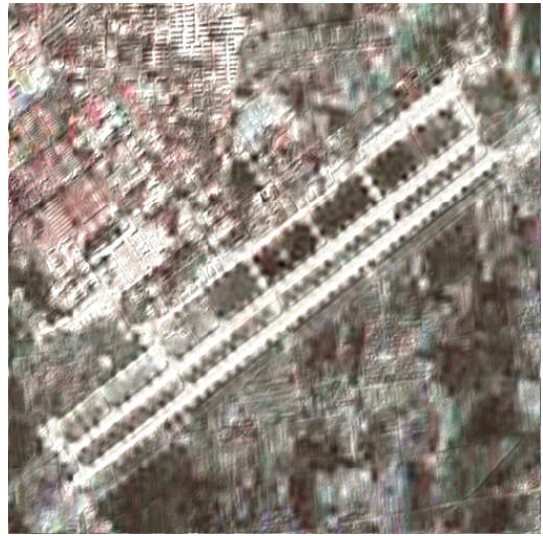

Fig..4b: HFM

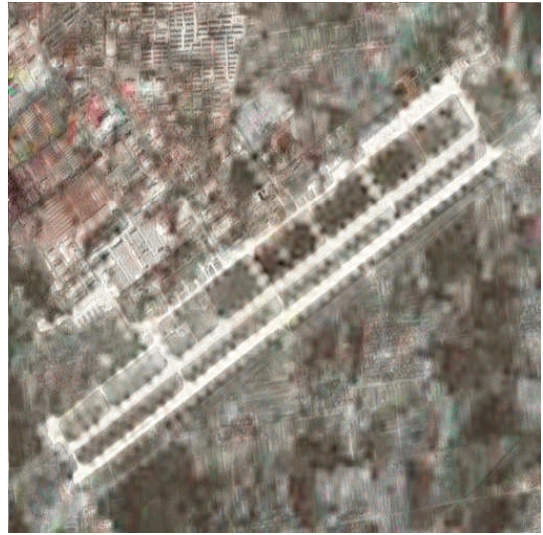

Fig..4c: HIS

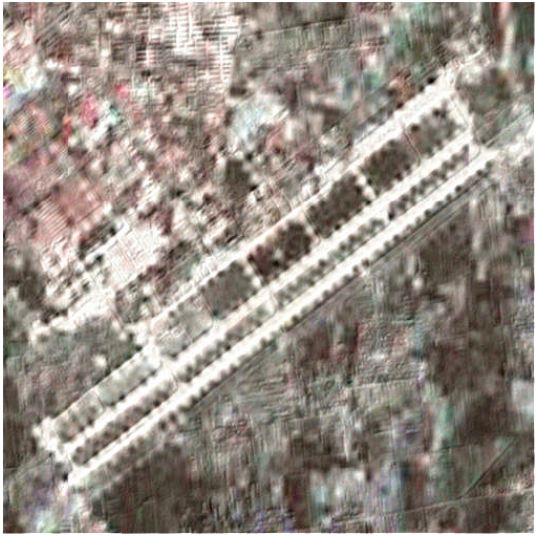

Fig..4a: HFA

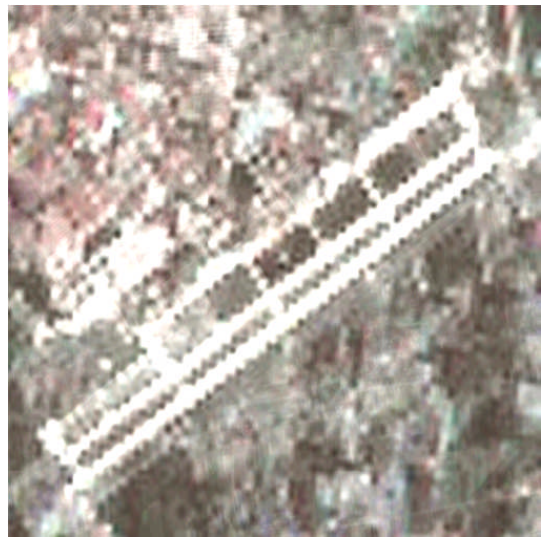

Fig.4d: PCA

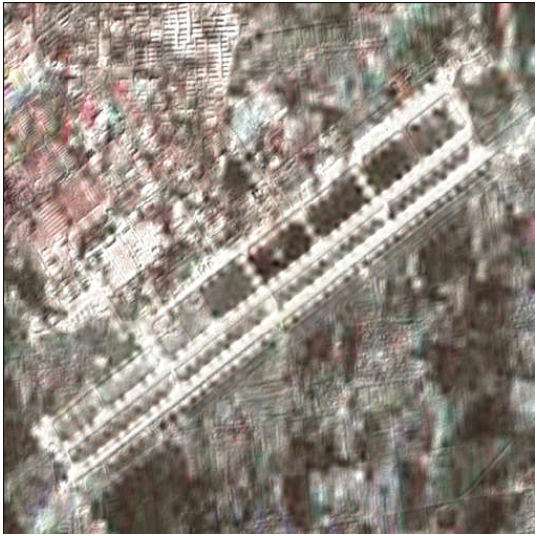

Fig..4e: RVS

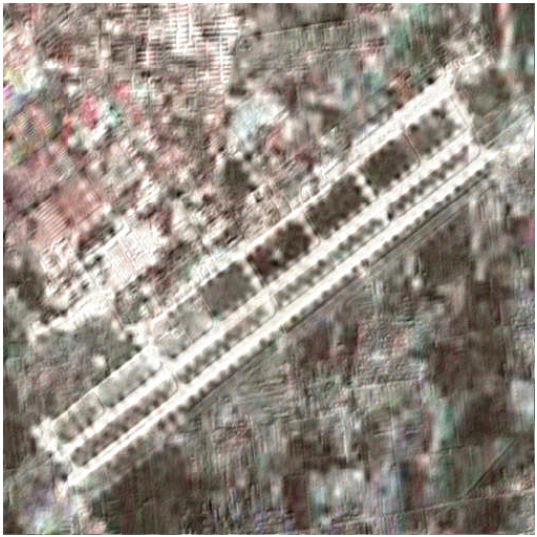

Fig.4f: SF

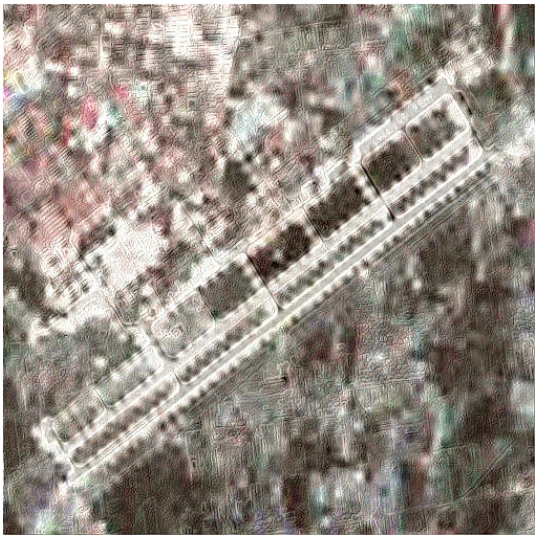

Fig..4g: EF

Fig.4: The Representation of Fused Images

## VI. CONCLUSION

This paper goes through the comparative studies undertaken by best different types of Image Fusion techniques based on pixel level as the following HFA, HFM, HIS and compares them with feature level fusion methods including PCA, SF and EF image fusion techniques. Experimental results with spatial and spectral quality matrices evaluation further show that the SF technique based on feature level fusion maintains the spectral integrity for MS image as well as improved as much as possible the spatial quality of the PAN image. The use of the SF based fusion technique is strongly recommended if the goal of the merging is to achieve the best representation of the spectral information of multispectral image and the spatial details of a high-resolution panchromatic image. Because it is based on Component Substitution fusion techniques coupled with a spatial domain filtering. It utilizes the statistical variable between the brightness values of the image bands to adjust the contribution of individual bands to the fusion results to reduce the color distortion.

The analytical technique of SG is much more useful for measuring the gradient than MG since the MG gave the smallest gradient results. The our proposed a approach HPDI gave the smallest different ratio between the image fusion methods, therefore, it is strongly recommended to use HPDI for measuring the spatial resolution because of its mathematical and more precision as quality indicator.

**Short Biodata of the Author**

Firouz Abdullah Al-Wassai. Received the B.Sc. degree in, Physics from University of Sana'a, Yemen, Sana'a, in 1993. The M.Sc.degree in, Physics from Bagdad University , Iraqe, in 2003, Research student.Ph.D in the department of computer science (S.R.T.M.U), Nanded, India.

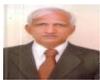Dr. N.V. Kalyankar, Principal, Yeshwant Mahvidyalaya, Nanded(India) completed M.Sc.(Physics) from Dr. B.A.M.U, Aurangabad. In 1980 he joined as a leturer in department of physics at Yeshwant Mahavidyalaya, Nanded. In 1984 he completed his DHE. He completed his Ph.D. from Dr.B.A.M.U. Aurangabad in 1995. From 2003 he is working as a Principal to till date in Yeshwant Mahavidyalaya, Nanded. He is also research guide for Physics and Computer Science in S.R.T.M.U, Nanded. 03 research students are successfully awarded Ph.D in Computer Science under his guidance. 12 research students are successfully awarded M.Phil in Computer Science under his guidance He is also worked on various boides in S.R.T.M.U, Nanded. He is also worked on various bodies is S.R.T.M.U, Nanded. He also published 34 research papers in various international/national journals. He is peer team member of NAAC (National Assessment and Accreditation Council, India ). He published a book entilteld "DBMS concepts and programming in Foxpro". He also get various educational wards in which "Best Principal" award from S.R.T.M.U, Nanded in 2009 and "Best Teacher" award from Govt. of Maharashtra, India in 2010. He is life member of Indian "Fellowship of Linnean Society of London(F.L.S.)" on 11 National Congress, Kolkata (India). He is also honored with November 2009.

Dr. Ali A. Al-Zuky. B.Sc Physics Mustansiriyah University, Baghdad , Iraq, 1990. M Sc. In1993 and Ph. D. in1998 from University of Baghdad, Iraq. He was supervision for 40 postgraduate students (MSc. & Ph.D.) in different fields (physics, computers and Computer Engineering and Medical Physics). He has More than 60 scientific papers published in scientific journals in several scientific conferences.